# COBRA: Contrastive Bi-Modal Representation Algorithm

Vishaal Udandarao[1*], Abhishek Maiti[1*], Deepak Srivatsav[1*], Suryatej Reddy Vyalla[1*]
Yifang Yin[2], Rajiv Ratn Shah[1]
[1]IIIT-Delhi, India; [2]NUS, Singapore
{vishaal16119,abhishek16005,deepak16030,suryatej16102,rajivratn}@iiitd.ac.in,yifang@comp.nus.edu.sg

**Abstract**

There are a wide range of applications that involve multi-modal data, such as cross-modal retrieval, visual question-answering and image captioning. Such applications are primarily dependent on aligned distributions of the different constituent modalities. Existing approaches generate latent embeddings for each modality in a joint fashion by representing them in a common manifold. However these joint embedding spaces fail to sufficiently reduce the modality gap, which affects the performance in downstream tasks. We hypothesize that these embeddings retain the *intra*-class relationships but are unable to preserve the *inter*-class dynamics. In this paper, we present a novel framework COBRA that aims to train two modalities (*i.e.*, image and text) in a joint fashion inspired by the Contrastive Predictive Coding (CPC) and Noise Contrastive Estimation (NCE) paradigms which preserve both inter-class and intra-class relationships. We have conducted extensive experiments on four diverse downstream tasks spanning across seven benchmark cross-modal datasets. The experimental results show that our proposed framework outperforms existing work by 1.09% - 22%, as it generates a robust and task agnostic joint-embedding space.

**CCS Concepts**

• **Computing methodologies** → **Learning latent representations**; • **Information systems** → **Multimedia information systems**.

**Keywords**

Joint embedding spaces, latent representations, contrastive learning, bi-modal data

## 1 Introduction

Systems built on multi-modal data have been shown to perform better than systems that solely use uni-modal data [7, 49]. Due to this fact, multi-modal data is widely used in and generated by different large-scale applications. These applications often utilize this multi-modal data for tasks such as information retrieval [11, 44], classification [48, 58], and question-answering [27, 35]. It is therefore important to represent such multi-modal data in a meaningful and interpretable fashion to enhance the performance of these large-scale applications. In this work, we focus on learning the joint cross-modal representations for images and text, but our proposed techniques can be easily extended to other modalities as well. Learning meaningful representations for multi-modal data is challenging because there exists a distributional shift between different modalities [18, 37]. The lack of consistency in representations across modalities further magnifies this shift [6]. Due to such difficulties, any similarity metric between the representations across modalities is intractable to compute [37]. The reduction of this distributional shift boils down to two challenges: (1) projecting the representations of data belonging to different modalities to a common manifold (also referred to as the joint embedding space), and (2) retaining their semantic relationship with other samples from the same class as well as different classes.

The need for a joint embedding space is emphasized by the inability of uni-modal representations to align well with each other. Over the last few years, literature [18, 29, 36] has been presented where the representations were modeled in the joint embedding space, but existing methods perform less satisfactorily as significant semantic gap still exists among the learnt representations from different modalities. We believe this is due to the fact that current cross-modal representation systems regularize the distance of pairs of representations of those data samples which belong to the same classes (but different modalities) but not of pairs of representations belonging to different classes (can be from the same or different modalities).

While current work [18, 36] has focused on conserving the semantic relationship between *intra* cross-modal data, *i.e.*, belonging to the same class, we surmise that along with this, preserving *inter* cross-modal interactions will help the model learn a more discriminatory boundary between different classes.

**Motivation:** We posit that preserving the relationship between representations of samples belonging to different classes, in a modality invariant fashion, can improve the quality of joint cross-modal embedding spaces. We formulate this hypothesis as it introduces a contrastive proximity mechanism between data belonging to different semantic classes. This distancing will allow the model to converge to a better generalizing decision boundary. Similar *contrastive learning paradigms* based on information gain have been performing very well in the self-supervised learning problem settings [19, 56, 59]. *To the best of our knowledge, we are the first to propose a method to learn joint cross-modal embeddings based on contrastive learning paradigms.*

**Contributions:** Our contributions are as follows:

- We propose a novel joint cross-modal embedding framework called COBRA (COntrastive Bi-modal Representation Algorithm) which represents the data across different modalities (text and image in this study) in a common manifold.
- We utilize a set of loss functions in a novel way, which jointly preserve not only the relationship between different *intra* cross-modal data samples but also preserve the relationship between *inter* cross-modal data samples (refer Figure 1).
- We empirically validate our model by achieving state-of-the-art results on four diverse downstream tasks: (1) cross-modal retrieval, (2) finegrained multi-modal sentiment classification,

---

*Equal contribution. Ordered Randomly.

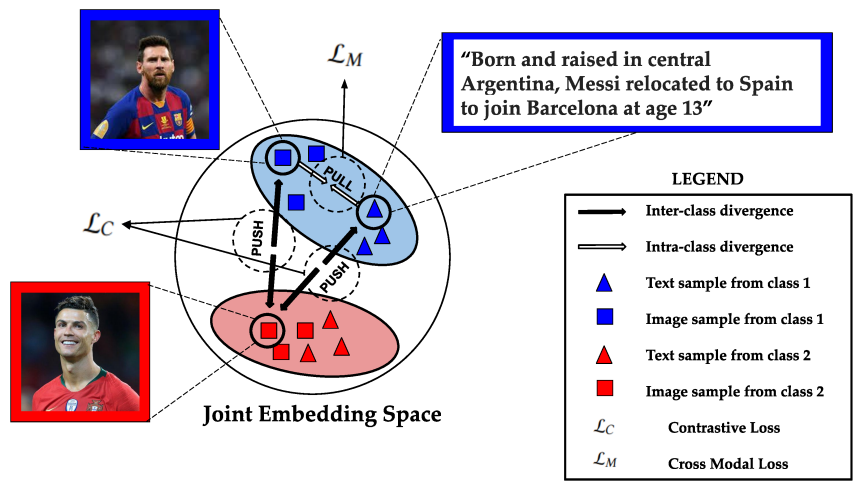

Figure 1: Visualization of the working of contrastive and cross modal losses. The contrastive and cross modal losses enforce divergence across samples of different classes but ensure that the samples of the same class are drawn together, regardless of their modality. This ascertains that the joint embedding space is both modality invariant and class discriminative.

(3) multi-modal disaster classification, and (4) multi-modal fake news detection.

The rest of the paper is structured as follows. Section 2 covers the adjoining work done in the field of cross-modal retrieval systems and multi-modal fusion along with an introduction to the *contrastive learning paradigms*, which forms the theoretical backbone of our approach. Section 3 presents our proposed methodology. Section 4 provides a description of the benchmark datasets and the evaluation metrics used. Section 5 concludes the paper.

## 2 Related Work

In this section, we discuss the topics that inspire the architecture and loss functions used in COBRA: Cross-modal Correlation, Multi-modal Fusion, Contrastive Learning Paradigms and Noise Contrastive Estimation. We further discuss the tasks used to evaluate COBRA: Cross-modal Retrieval, Multi-modal Fake News Detection, Multi-modal Sentiment Classification, and Multi-modal Disaster Classification.

### 2.1 Cross-modal Correlation

The initial attempts at using deep architectures for establishing cross-modal correlation relationships were inspired from the traditional methods [14, 45]. Andrew *et al.* [4] proposed Deep Canonical Correlation Analysis (DCCA) based on the older Canonical Correlation Analysis (CCA) [17]. DCCA works for only two modalities and is able to generate highly correlated (linear) projections. Kan *et al.* [22] were one of the first to work in the supervised setting and proposed a network that learns a common discriminative subspace using Fisher's criterion in a neural network. Cross Media Multiple Deep Network (CMDN) [36] and Cross Modal Correlation Learning (CCL) [38] are two networks that utilize the inter and intra modality correlation and learn the common representations of multi-modal data.

### 2.2 Multi-modal Fusion

Significant amount of work in the domain of multimedia research has been based on fusion techniques for datasets of multiple modalities. The type of fusion affects the dynamics of the features produced. Early fusion techniques that are based on simple concatenation [41, 65] do not capture the intra modal relations well. Late fusion techniques [21, 33] on the other hand prioritize intra modal learning abilities compromising on cross-modal differentiability. This is because these models make decisions on a weighted average score of individual modality features. To solve both these limitations, Mai *et al.* [29] proposed an adversarial representation learning and graph fusion network for multi-modal fusion. They employed adversarial techniques to learn a modality-invariant embedding space and used a hierarchical graph neural network to capture multi-modal interactions. Fusion networks have also shown great performance in application specific tasks. Ding *et al.* [10] proposed a fusion based DNN for predicting popularity on social media. Further, Hong *et al.* [16] proposed a deep fusion network for the task of image completion. However, literature suggests that cross modal tasks benefit more from learning a joint embedding space than employing multi-modal fusion techniques [7].

### 2.3 Contrastive Learning Paradigms

Contrastive Learning techniques have gained popularity recently because of their success in unsupervised settings. Oord *et al.* [59] were one of the first to propose a Contrastive Predictive Coding (CPC) technique that could generate useful representations from high dimensional data universally in an unsupervised fashion. Chen *et al.* [19] improved upon this and proposed a CPCv2 that obtained better results on classification and object detection tasks. Further, Tian *et al.* [56] developed a compact representation that maximized mutual information between different views of the same scene and hence improved performance on image and video unsupervised learning tasks. Chen *et al.* [8] proposed a Simple Framework for Contrastive Learning of Visual Representations (SimCLR) that eliminated the requirement of specialized architectures or memory banks for contrastive taks and also gave state-of-the-art results on self-supervised classification tasks. All these techniques proposed so far have been employed only for single modality tasks. Arora *et al.* [5] conducted an extensive theoretical study on the efficacy of using these contrastive algorithms on several downstream tasks and showed that these algorithms lead to learning richer latent embeddings.

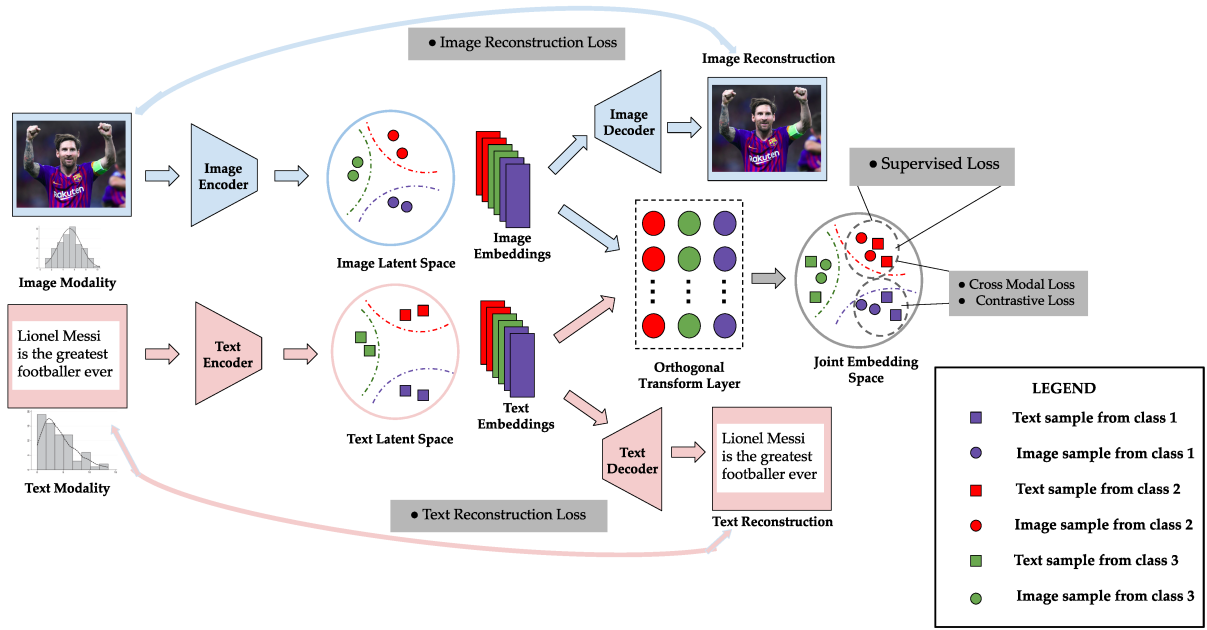

**Figure 2: The general architecture of our proposed COBRA model. The different shapes are present to help visualize the structure of the joint embedding space with respect to the orthogonal projections of both image and text samples. The losses of our model are highlighted in the grey boxes.**

### 2.4 Noise Contrastive Estimation

Almost all the contrastive paradigms detailed in the previous section rely on some sort of negative sampling techniques to reduce the complexity of computing representations. Gutmann *et al.* [15] first proposed NCE as a technique to model discriminations between a true statistical distribution and an artificial noise distribution. They used non-linear regression to estimate true distributions via contrastive estimation against randomly drawn noise samples. Since then, several applications have made use of NCE to estimate sample distributions. Mnih *et al.* [32] leveraged NCE to learn word embeddings over large vocabulary spaces in a scalable and efficient manner. They showed that this method achieved state-of-the-art results with minimal training time. Rao *et al.* [43] utilized an NCE based loss to achieve state-of-the-art results for the problem of answer-selection. They did so by modelling the problem as a point-wise classification problem and optimizing the model trained using NCE. More recently, Amrani *et al.* [3] applied NCE to learn robust density estimates for different modalities. Through this approach they were able to perform well on diverse downstream tasks including visual question-answering and text-to-video retrieval.

### 2.5 Cross-modal Retrieval

We focus on the literature based on generating joint embeddings for cross modal retrieval. Wu *et al.* [64] proposed a technique to generate embeddings that preserved the semantic structure of the labels in the data. Mithun *et al.* [30] discussed a framework that utilized multi-modal cues from videos for video-text retrieval. Mithun *et al.* [31] leveraged web images and tags to learn a visual-semantic joint embedding . Further, Hu *et al.* [18] proposed a scalable autoencoder based architecture to learn a smooth and pure label representation space (shared across modalities) for any number of input modalities.

### 2.6 Multi-modal Fake News Detection

Deep neural networks have shown great potential in detecting multi-modal fake news. Wang *et al.* [62] discussed a model that learnt event invariant feature representations across modalities using an adversarial network. Khattar *et al.* [23] proposed Multi-modal Variational Autoencoder (MVAE) network that employed a bi-modal variational autoencoder for fake news classification. Singhal *et al.* [52] proposed an architecture SpotFake+ that leveraged pre-trained language transformers and convolutional neural networks for detecting fake news.

### 2.7 Multi-modal Sentiment Classification

Multi-modal sentiment classification is a promising area of research and there have been many attempts at using deep neural networks for this task. Chen *et al.* [68] developed Tensor Fusion Networks, an end-to-end model which learns both intra-modality and inter-modality dynamics for the task of sentiment classification. Pham et al. [40] proposed Seq2Seq2Sentiment, an unsupervised method for learning joint multi-modal representations using sequence to sequence models. Wang et al. [63] discussed a new fusion method TransModality using transformers in an end-to-end fashion for multi-modal sentiment analysis.

## 2.8 Multi-modal Disaster Classification

Gautam et al. [13] developed a novel decision diffusion technique on the CrisisMMD dataset [2, 34] to classify disaster related data into informative and non-informative categories using image and text uni-modal models. Agarwal et al. [1] proposed Multimodal Emergency Management Information System (MEMIS) that leverages both visual and textual features on the same dataset. Their system outperforms all other existing uni-modal methods.

## 3 Methodology

In this section, we first explain the formulation of our problem statement in terms of the data we use. We then introduce and explain the architecture of our model, along with the loss functions used. We finally explain our optimization and training strategy.

## 3.1 Problem Formulation

Let us assume that we have two modalities, *i.e.* text and image, we denote the $j$-th image sample as $x_I^j \in \mathbb{R}^{d_I}$ and the $j$-th text sample as $x_T^j \in \mathbb{R}^{d_T}$. Here, $d_I$ and $d_T$ represent the dimensionality of the image and text samples respectively. We denote the image dataset as $X_I = \{x_I^0, x_I^1, ..., x_I^{n_I-1}\}$ and the text dataset as $X_T = \{x_T^0, x_T^1, ..., x_T^{n_T-1}\}$, where $n_I$ and $n_T$ denote the total number of data samples in the image and text datasets respectively. The corresponding labels for the image and text modalities are represented as follows: $Y_I = [y_I^0, y_I^1, ..., y_I^{n_I-1}]$ and $Y_T = [y_T^0, y_T^1, ..., y_T^{n_T-1}]$. Assuming there are $C$ distinct semantic classes in our multi-modal dataset, the labels are: $y_I^{j_I}, y_T^{j_T} \in \{0, 1, ..., C-1\} \forall j_I \in \{0, 1, ..., n_I-1\}, j_T \in \{0, 1, ..., n_T-1\}$.

## 3.2 Model Architecture

The overall architecture for our model is given in Figure 2. Our goal is to represent the data in a common manifold, such that the class-wise representations are modality invariant and discriminatory. To this end, we use an autoencoder for each modality to generate representations that are high fidelity in nature. We utilize an orthogonal transform layer, which takes as input the hidden space representations from the encoders of each modality, and projects these representations into a joint space that is modality invariant and discriminates between classes well.

We denote the encoded representation as $z_j^i = f_j(x_j^i, \Theta_j)$ and the reconstructed sample as $\hat{x}_j^i = g_j(z_j^i, \Phi_j)$ where $i \in \{0, n_T-1\}$ and $i \in \{0, n_I-1\}$ for text and image respectively, and where $j \in \{T, I\}$ for text and image respectively. $f_j$ denotes the encoder of the $j$-th modality parameterised by $\Theta_j$. Similarly $g_j$ denotes the decoder of the $j$-th modality parameterised by $\Phi_j$. Given the representations $z_T^i$ and $z_I^i$, which have dimensions $Z_T$ and $Z_I$, we project the representations to a joint subspace such that the representation of each semantic class is orthogonal to each other [18]. We call these projections $O_T^i$ and $O_I^i$, both of which have dimension $Z$.

We define the loss function in COBRA as a weighted sum of the reconstruction loss, cross-modal loss, supervised loss and contrastive loss, the details of which are introduced below. To preserve the inter-class dynamics, we innovatively introduce the *Contrastive Loss* that has never been used in representing multi-modal data.

### 3.2.1 Reconstruction Loss

Reconstruction loss has been used in the autoencoder. Given the decoder output $\hat{x}_i^j$ and the input $x_i^j$, we define the reconstruction loss shown in Eq. 1 as:

$$\mathcal{L}_R = \sum_{i \in \{I,T\}} \sum_{j=0}^{n_i-1} \left\| \hat{x}_i^j - x_i^j \right\|_2^2 \quad (1)$$

### 3.2.2 Cross-Modal Loss

The projected representations $O_I^j$ and $O_T^j$ align class representations within each modality. The cross-modal loss aims to align representations of the same class across different modalities. Given the projected representations $O_I^j$ and $O_T^i$, we define the cross-modal loss shown in Eq. 2 as:

$$\mathcal{L}_M = \sum_{j=0}^{\min\{n_T,n_I\}-1} \left\| O_T^j - O_I^j \right\|_2^2 \quad (2)$$

We use the min function because the dataset may not have equal text and image samples. We only take those pairs in which the corresponding text and image samples are present.

### 3.2.3 Supervised Loss

As we try to model an orthogonal latent space having the joint embeddings, we utilize the one-hot labels of the data samples to reinforce those samples belonging to the same class but different modalities to be grouped together in the same subspace. Let $\hat{y}_i^j$ be the one-hot encoded label for the $j$-th sample of the $i$-th modality, and $O_i^j$ be the projected representation, we then define the supervised loss shown in Eq. 3 as:

$$\mathcal{L}_S = \sum_{i \in \{I,T\}} \sum_{j=0}^{n_i-1} \left\| O_i^j - \hat{y}_i^j \right\|_2^2 \quad (3)$$

### 3.2.4 Contrastive Loss

As stated in recent literature [5, 56, 57], to implement the contrastive loss [15, 54], the definitions of positive samples and negative samples of representations are of utmost importance. We will first define the positive and negative samples pertaining to our model. Given the projected representations $O_I^i$ and $O_T^i$, a positive pair is defined as the representations of data samples belonging to the same modality and class. A negative pair is defined as the representations of two data samples belonging to same or different modality of different classes. To define the contrastive loss, a scoring function is required, which yields high values for positive samples and low values for negative values. Here we define the scoring function by taking the dot product of the representations in the joint embedding space. Following several recent works [8, 19, 24, 59], our loss function enforces the model to select the positive sample from a fixed sized set $S = \{p, n_1, n_2, ..., n_N\}$ containing one positive and $N$ negative samples. Thereafter we formulate our contrastive loss shown in Eq. 4 as:

$$\mathcal{L}_C = -\mathbb{E}_S \left[ \log \frac{a^T p}{a^T p + \sum_{i=1}^{N} a^T n_i} \right] \quad (4)$$

where $a$ is the anchor point, $p$ is its corresponding positive sample, $\mathbb{E}$ is an expectation operator over all possible permutations of $S$ and $n_i$ iterates over all the negative samples. The anchor, positive and negative samples are randomly drawn from each mini-batch. We minimize the above expectation running over all samples. Since fetching negative samples from the entire dataset is computationally infeasible, we sample the negative points only from each mini-batch locally.

Since, we sample only a finite sized set of negative samples, the model can miss out on characteristics of the distribution of the joint embeddings. To avoid this, we implement another loss called the Noise Contrastive Estimation (NCE) [15] loss, which is an effective method for estimating unnormalized models. NCE helps to model the distribution of the negative samples by leveraging a proxy noise distribution. It does so by estimating the probability of a sample coming from a joint distribution rather than it coming from a noise distribution. The noise distribution is assumed to be a uniform distribution. Denoting the joint distribution of positive samples as $p_J$, the noise distribution as $p_N$, the anchor sample as $a$ and every other sample (can be either positive or negative) as $s$, the probability of data sample $s$ coming from the joint distribution $p_J$ is:

$$P(X = 1|s; a) = \frac{p_J(s|a)}{p_J(s|a) + N p_N(s|a)} \quad (5)$$

where $N$ is the number of samples from the noise distribution. Instead of using Eq. 4, now we can estimate the contrastive loss more accurately based on Eq. 6 as follows:

$$\mathcal{L}_C = -\mathbb{E}_a \{ \mathbb{E}_{s \sim p_J(\bullet|s)} [[P(X = 1|s; a)]] \\ + N \times \mathbb{E}_{s \sim p_N(\bullet|s)} [1 - P(X = 1|s; a)] \} \quad (6)$$

where $\mathbb{E}_a$ is an expectation over all possible anchor samples, $\mathbb{E}_{s \sim p_J}$ is an expectation over all possible positive samples (corresponding to anchor $a$) from the joint distribution $p_J$, and $\mathbb{E}_{s \sim p_N}$ is an expectation over all samples from the noise distribution $p_N$.

### 3.3 Optimization and Training Strategy

The overall loss of our network is defined to be a weighted sum of the reconstruction loss, cross-modal loss, supervised loss and contrastive loss. The weights are treated as hyperparameters.

$$\mathcal{L} = \lambda_R \mathcal{L}_R + \lambda_S \mathcal{L}_S + \lambda_M \mathcal{L}_M + \lambda_C \mathcal{L}_C \quad (7)$$

The objective function in Eq. 7 is optimized using stochastic gradient descent. The loss is summed over all modalities, and the corresponding gradient is propagated through all the components in the model. The optimization process of our proposed network is illustrated in Algorithm 1. We adopted the PyTorch framework for implementation, and trained our models for 200 epochs on an Nvidia GTX 1050 GPU[1].

---
[1]Code available at https://github.com/ovshake/cobra

---

**Algorithm 1:** Flow of the COBRA algorithm

**Input** : The image training set $X_I$, the text training set $X_T$, the image label set $Y_I$, the text label set $Y_T$, dimensionality of the joint embedding space $Z$, image batch size $b_I$, text batch size $b_T$, learning rate $\eta$, hyperparameters $\lambda_M, \lambda_C, \lambda_S, \lambda_R$, number of training epochs $N$ and number of iterations (batch count) per epoch $B$

**Output**: The optimal encoder weights $\Theta_I, \Theta_T$ and optimal decoder weights $\Phi_I, \Phi_T$

1 Initialize $\Theta_I, \Theta_T, \Phi_I, \Phi_T$ randomly
2 **for** $i=1,2,...,N$ **do**
3     **for** $b=1,2,...,B$ **do**
4        Sample a random text minibatch $m_T$ of size $b_T$
5        Sample a random image minibatch $m_I$ of size $b_I$
6        Compute the image and text encoded latent representations $z_I$ and $z_T$
7        Compute the image and text orthogonal projections $O_I$ and $O_T$
8        Compute the image and text reconstructions $\hat{x}_I$ and $\hat{x}_T$
9        Compute the losses: $\mathcal{L}_R$ (Eq. 1), $\mathcal{L}_M$ (Eq. 2), $\mathcal{L}_S$ (Eq. 3), and $\mathcal{L}_C$ (Eq. 4, 6)
10       Compute total loss (Eq. 7) : $\mathcal{L} = \lambda_S \mathcal{L}_S + \lambda_R \mathcal{L}_R + \lambda_M \mathcal{L}_M + \lambda_C \mathcal{L}_C$
11       Update model weights using a SGD update rule:
12       $\Theta_I \leftarrow \Theta_I - \eta \frac{\partial \mathcal{L}}{\partial \Theta_I}; \Theta_T \leftarrow \Theta_T - \eta \frac{\partial \mathcal{L}}{\partial \Theta_T}$
13       $\Phi_I \leftarrow \Phi_I - \eta \frac{\partial \mathcal{L}}{\partial \Phi_I}; \Phi_T \leftarrow \Phi_T - \eta \frac{\partial \mathcal{L}}{\partial \Phi_T}$

## 4 Experiments

To evaluate our proposed method, we test our model on four different tasks, namely, cross-modal retrieval, multi-modal fake news detection, multi-modal sentiment classification, and multi-modal disaster classification. We compare the performance of our model against state-of-the-art models of corresponding tasks.

In the following sections, we describe the datasets and evaluation metrics adopted, followed by the results achieved on each downstream task mentioned above.

### 4.1 Cross-Modal Retrieval

In the task of cross-modal retrieval, we use COBRA to retrieve an image given a text query, or a text sample given an image query.

#### 4.1.1 Datasets

For the cross-modal retrieval task, we utilize four different datasets. For Wikipedia [46], MS-COCO [26], and NUS-Wide 10k [9] datasets, we convert the images into 4096-dimensional feature vectors using the fc7 layer of VGGnet [51]. In the Wikipedia and MS-COCO dataset, we convert the texts into 300-dimensional feature vectors using Doc2Vec [25]. For the NUS-Wide 10k dataset, we convert the text into 1000-dimensional Bag of Words feature vectors. The PKU-XMedia dataset [39, 69] contains texts represented as 3000-dimensional Bag of Words feature vectors and images represented

Table 1: Performance (mAP) on the Wikipedia Dataset

| Method | Image → Text | Text → Image | Average |
| --- | --- | --- | --- |
| MCCA [47] | 0.202 | 0.189 | 0.195 |
| ml-CCA [42] | 0.388 | 0.356 | 0.372 |
| DDCAE [61] | 0.308 | 0.290 | 0.299 |
| JRL [70] | 0.343 | 0.376 | 0.330 |
| ACMR [60] | 0.479 | 0.426 | 0.452 |
| CMDN [36] | 0.487 | 0.427 | 0.457 |
| CCL [38] | 0.504 | 0.457 | 0.481 |
| D-SCMR [72] | 0.521 | 0.478 | 0.499 |
| SDML [18] | 0.522 | 0.488 | 0.505 |
| DAML [66] | 0.559 | 0.481 | 0.520 |
| **COBRA** | **0.742** | **0.739** | **0.740** |

Table 2: Performance (mAP) on the MS-COCO Dataset

| Method | Image → Text | Text → Image | Average |
| --- | --- | --- | --- |
| MCCA [47] | 0.646 | 0.640 | 0.643 |
| ml-CCA [42] | 0.667 | 0.661 | 0.664 |
| DDCAE [61] | 0.412 | 0.411 | 0.411 |
| ACMR [60] | 0.692 | 0.687 | 0.690 |
| DCCA [4] | 0.415 | 0.414 | 0.415 |
| GSS-SL [71] | 0.707 | 0.702 | 0.705 |
| SDML [18] | 0.827 | 0.818 | 0.823 |
| **COBRA** | **0.854** | **0.853** | **0.853** |

Table 3: Performance (mAP) on the PKU-XMedia Dataset

| Method | Image → Text | Text → Image | Average |
| --- | --- | --- | --- |
| MCCA [47] | 0.620 | 0.616 | 0.618 |
| DDCAE [61] | 0.868 | 0.878 | 0.873 |
| JRL [70] | 0.770 | 0.788 | 0.779 |
| ACMR [60] | 0.882 | 0.885 | 0.883 |
| CMDN [36] | 0.485 | 0.516 | 0.501 |
| DCCA [4] | 0.869 | 0.871 | 0.870 |
| GSS-SL [71] | 0.875 | 0.878 | 0.876 |
| SDML [18] | 0.899 | 0.917 | 0.908 |
| **COBRA** | **0.945** | **0.941** | **0.943** |

Table 4: Performance (mAP) on the NUS-Wide 10k dataset

| Method | Image → Text | Text → Image | Average |
| --- | --- | --- | --- |
| MCCA [47] | 0.448 | 0.462 | 0.455 |
| DDCAE [61] | 0.511 | 0.540 | 0.525 |
| JRL [70] | 0.586 | 0.598 | 0.592 |
| ACMR [60] | 0.588 | 0.599 | 0.593 |
| CMDN [36] | 0.492 | 0.515 | 0.504 |
| CCL [38] | 0.506 | 0.535 | 0.521 |
| DCCA [4] | 0.532 | 0.549 | 0.540 |
| SDML [18] | 0.55 | 0.505 | 0.527 |
| DAML [66] | 0.512 | 0.534 | 0.523 |
| **COBRA** | **0.703** | **0.701** | **0.702** |

as 4096-dimensional feature vectors, generated using the fc7 layer of VGGnet [51].

- The Wikipedia dataset [46] contains 2866 text-image pairs, divided into 10 semantic classes, such as warfare, art & architecture and media. We use a training, validation and test set of 2173, 231 and 462 text-image pairs [46] respectively.
- The PKU-Xmedia dataset [39, 69] contains 5000 text-image pairs, divided into 20 semantic classes. We use a training, validation and test set of 4000, 500 and 500 text-image pairs [39, 69] respectively.
- The MS-COCO dataset [26] contains 82079 text-image pairs, divided into 80 semantic classes. We use a training, validation and test set of 57455, 14624 and 10000 text-image pairs [18] respectively.
- The NUS-Wide 10k dataset [9] contains 10000 text-image pairs, divided into 10 semantic classes. We use a training, validation and test set of 8000, 1000 and 1000 text-image pairs [60] respectively.

### 4.1.2 Evaluation Metrics
We compare our performance against state-of-the-art models based on Mean Average Precision (mAP). For a fair comparison, we ensure that we use the same features across models.

### 4.1.3 Results
We report the highest mAP for Text to Image (TTI) and Image to Text (ITT) retrieval on all four datasets.

From the t-SNE [28] plot for Wikipedia given in Figure 3a, we observe that COBRA is able to effectively form joint embeddings for different classes across modalities, resulting in superior performances across the aforementioned datasets.

We achieve a 22% improvement over the previous state-of-the-art (DAML [66]) on the Wikipedia dataset (Table 1). We achieve a 3% improvement over the previous state-of-the-art (SDML [18]) on the MS-COCO dataset (Table 2). We achieve a 3.5% improvement over the previous state-of-the-art (SDML [18]) on the PKU-XMedia dataset (Table 3). We achieve a 10.9% improvement over the previous state-of-the-art (ACMR [60]) on the NUS-Wide 10k dataset (Table 4).

## 4.2 Multi-modal Fake News Detection

In the task of multi-modal fake news detection, we use COBRA to determine whether a given bi-modal query (text and image) corresponds to a real or fake news sample.

### 4.2.1 Datasets
For the multi-modal fake news detection task, we utilize the Fake-NewsNet Repository [50]. This repository contains two datasets, namely, Politifact and Gossipcop. These datasets contain news content, social context, and dynamic information. We pre-process the data similar to Spotfake+ [52]. For both datasets, we convert images into 4096-dimensional feature vectors using VGGnet [51], and we convert texts into 38400-dimensional feature vectors using XLNet

COBRA: Contrastive Bi-Modal Representation AlgorithmTable 5: Accuracy on the FakeNewsNet dataset

| Method | Politifact (%) | Gossipcop (%) |
| --- | --- | --- |
| EANN [62] | 74 | 86 |
| MVAE [23] | 67.3 | 77.5 |
| SpotFake [53] | 72.1 | 80.7 |
| SpotFake+ [52] | 84.6 | 85.6 |
| **COBRA** | **86** | **86.7** |

[67]. Each dataset contains two semantic classes, namely, Real and Fake.

- The Politifact dataset contains 1056 text-image pairs. We get 321 Real and 164 Fake text-image pairs after pre-processing. We use a training, validation and test set of 381, 50 and 54 text-image pairs [52] respectively.
- The Gossipcop dataset contains 22140 text-image pairs. We get 10259 Real and 2581 Fake text-image pairs after pre-processing. We use a training, validation and test set of 10010, 1830 and 1000 text-image pairs [52] respectively.

#### 4.2.2 Evaluation metrics

We compare our performance against existing state-of-the-art models based on number of correctly classified queries (accuracy). For the purpose of our evaluation, we ensure that we use the same features that were used across other existing state-of-the-art models. To visualize the purity of the joint embedding space for different classes and modality samples, we plot the joint embeddings of COBRA trained on both the Gossipcop and Poltifact datasets. We plot the embeddings (Figure 3b and 3c) by employing the t-SNE [28] transformation to reduce the high dimensional joint embeddings ($O_I$ and $O_T$) to 3 dimensional data points. The figures clearly exhibit the high discrimination between samples of different classes in the joint embedding space. This provides further empirical validation for the high class divergence across the joint embedding space, irrespective of the modalities of the data points.

#### 4.2.3 Results

We achieve a 1.4% and a 1.1% improvement over the previous state-of-the-art (SpotFake+ [52]) on the Politifact and Gossipcop dataset respectively (Table 5). On observing the t-SNE plots in Figure 3, we discern a high intra-class variability in the Gossipcop dataset. We believe that there is only a small improvement because of the high class imbalance in these two datasets.

### 4.3 Multi-modal Fine-grained Sentiment Classification

In the task of multi-modal fine-grained sentiment classification, we use COBRA to perform ten tasks of classifying a given bi-modal query (text and image) into a sentiment category.

#### 4.3.1 Datasets

For the multi-modal fine-grained sentiment classification task, we analyze the performance of our model on the MeTooMA dataset [12]. This dataset contains 9973 tweets that have been manually annotated into 10 classes, namely, text only informative and image only informative (Relevance), Support, Opposition and Neither (Stance), Directed Hate and Generalized Hate (Hate Speech), Allegation, Refutation and Justification (Dialogue acts), and sarcasm. We convert the images into 4096-dimensional feature vectors using the fc7 layer of VGGnet [51]. We convert the texts into 300-dimensional feature vectors using Doc2Vec [25]. We use a training, validation and test set of 4500, 1000 and 1000 text-image pairs respectively, across all models that we test.

#### 4.3.2 Evaluation Metrics

We report the number of correctly classified queries (accuracy). To the best of our knowledge, we are the first to test a multi-modal classification model on this dataset. To this end, we evaluate our model against a Text-only and Image-only baseline, and Early Fusion. For the baselines, we use a Fully Connected network[2].

#### 4.3.3 Results

We obtain an average classification accuracy of 88.32% across all classes on the MeTooMA Dataset. This is a 1.2% improvement over Early Fusion (Table 6). We observe a low increase in Text only and Image only informative tasks due to the fact that 53.2% of our training data had text-image pairs with conflicting labels, *i.e.*, from a given text-image pair, the text may be labelled as "relevant" whereas the corresponding image may be labelled as "irrelevant". Furthermore, for classes under the Hate Speech, Sarcasm, and Dialogue Acts categories, we observe that there are less than 600 samples for each class. In categories such as Stance, where the 'Support' class has over 3000 samples, we observe much larger improvements in performance.

### 4.4 Multi-modal Disaster Classification

In the task of multi-modal disaster classification, we use COBRA to perform three classification tasks given a bi-modal (text and image) query. The classification tasks are further explained in the dataset section as follows.

#### 4.4.1 Datasets

For the multi-modal disaster classification task, we utilize the CrisisMMD dataset [2, 34]. It consists of 16058 tweets and 18082 images that were collected during natural disasters. There are 3 classification tasks that can be performed on this dataset —

- Informative or Non-Informative classification – this represents whether or not a particular text-image pair from a tweet is informative.
- Humanitarian Categories classification – this includes classes such as affected individuals, vehicle damage, missing or found people, and infrastructure or utility damage. This is once again done for a particular text-image pair from a tweet.
- Damage severity assessment – this includes classes such as severe damage, mild damage and little or no damage. This is once again done for a particular text-image pair from a tweet.

We convert the images into 4096-dimensional feature vectors using the fc7 layer of VGGnet [51]. We convert the texts into 300-dimensional feature vectors using Doc2vec [25]. We use a training set of 2000 text-image pairs, a validation set of 793 text-image pairs for the first 2 classification tasks, a validation set of size 697 for the third classification task, and a test set of 500 text-image pairs.

---
[2]Architectural details can be found in the supplementary material

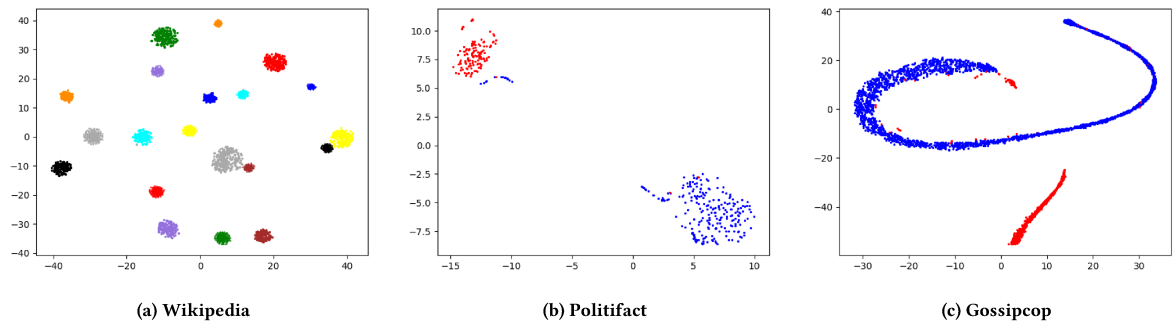

(a) Wikipedia  (b) Politifact  (c) Gossipcop

Figure 3: t-SNE visualizations of the joint embedding spaces of the models trained on Gossipcop, Politifact and Wikipedia datasets. The different colours correspond to the various class labels in the dataset.

Table 6: Accuracy on the MeTooMA Dataset

| Label | **COBRA** (%) | Text-only baseline (%) | Image-only baseline (%) | Early Fusion (%) |
|---|---|---|---|---|
| Text only informative | **73.77** | 73.43 | 63.39 | 72.15 |
| Image only informative | 67.36 | 63.21 | **67.74** | 66.97 |
| Directed Hate | **96.43** | 95.12 | 94.67 | 95.85 |
| Generalized Hate | **97.77** | 96.19 | 95.89 | 96.88 |
| Sarcasm | **98.55** | 96.94 | 96.45 | 97.16 |
| Allegation | **93.75** | 92.67 | 92.40 | 93.19 |
| Justification | **98.44** | 96.23 | 95.66 | 97.34 |
| Refutation | **98.54** | 96.90 | 96.81 | 97.37 |
| Support | **66.29** | 61.60 | 59.93 | 63.28 |
| Opposition | **92.3** | 90.1 | 89.5 | 91.1 |
| Average | **88.32** | 86.23 | 85.24 | 87.12 |

#### 4.4.2 Evaluation Metrics
We compare our performance against existing state-of-the-art models based on number of correctly classified queries (accuracy).

#### 4.4.3 Results
We obtain the following results on the three tasks (Table 7) —

- Informative or Non-Informative classification — we obtain an accuracy of 93.49% on this task, which is a 1.09 % improvement over Agarwal et al. [1].
- Humanitarian Categories — we obtain an accuracy of 42.25%, which is a 5.45 % improvement over Agarwal et al. [1].
- Damage severity assessment — we obtain an accuracy of 64.58%, which is an 8.38% improvement over Agarwal et al. [1].

We believe this improvement is achieved because of the good quality of the representations obtained from COBRA[3].

## 5 Conclusion

In this paper, we propose a novel approach (COBRA) to jointly learn bi-modal representations in an orthogonal space. We show that our proposed method learns better representations which allows the model to generalize the decision boundary in a much more robust fashion. This enables us to achieve state-of-the-art results on four downstream tasks. The representations learnt are high-fidelity in nature, containing sufficient information for reconstruction as well as tasks such as retrieval and classification. Different from other models, COBRA, along with preserving the *intra*-class relationship of samples in the embedding space, also preserves the *inter*-class relationship using a Contrastive Learning Paradigm called Noise Contrastive Estimation (NCE). This ensures that the samples belonging to the same class are clustered together, and that the distance between clusters of samples belonging to different classes (irrespective of the modality) is maximized in the joint embedding space. As for the future work, we attempt to extend our method to a self-supervised/semi-supervised problem setting.

[3]The t-SNE plots which further strengthen our claim can be found in the supplementary material

Table 7: Accuracy on the CrisisMMD Dataset. The labels are: *I* - 'Informativeness', *HC* - 'Humanitarian Categories', *DS* - 'Disaster Severity'

| Model | I (%) | HC (%) | DS (%) |
|---|---|---|---|
| Gautam et al. [13] | 73.57 | - | - |
| Agarwal et al. [1] — FastText [20] | 88.39 | 34.4 | 56.20 |
| Agarwal et al. [1] — Inception [55] | 92.40 | 37.0 | 43.40 |
| **COBRA** | **93.49** | **42.45** | **64.58** |

## References

[1] Mansi Agarwal, Maitree Leekha, Ramit Sawhney, Rajiv Ratn Shah, Rajesh Kumar Yadav, and Dinesh Kumar Vishwakarma. 2020. MEMIS: Multimodal Emergency Management Information System. In *Advances in Information Retrieval*, Joemon M. Jose, Emine Yilmaz, João Magalhães, Pablo Castells, Nicola Ferro, Mário J. Silva, and Flávio Martins (Eds.). Springer International Publishing, Cham, 479–494.

[2] Firoj Alam, Ferda Ofli, and Muhammad Imran. 2018. CrisisMMD: Multimodal Twitter Datasets from Natural Disasters. In *Proceedings of the 12th International AAAI Conference on Web and Social Media (ICWSM)* (USA, 23-28).

[3] Elad Amrani, Rami Ben-Ari, Daniel Rotman, and Alex Bronstein. 2020. Noise Estimation Using Density Estimation for Self-Supervised Multimodal Learning. *arXiv preprint arXiv:2003.03186* (2020).

[4] Galen Andrew, Raman Arora, Jeff Bilmes, and Karen Livescu. 2013. Deep Canonical Correlation Analysis. In *Proceedings of the 30th International Conference on Machine Learning (Proceedings of Machine Learning Research)*, Sanjoy Dasgupta and David McAllester (Eds.), Vol. 28. PMLR, Atlanta, Georgia, USA, 1247–1255. http://proceedings.mlr.press/v28/andrew13.html

[5] Sanjeev Arora, Hrishikesh Khandeparkar, Mikhail Khodak, Orestis Plevrakis, and Nikunj Saunshi. 2019. A theoretical analysis of contrastive unsupervised representation learning. *arXiv preprint arXiv:1902.09229* (2019).

[6] Devanshu Arya, Stevan Rudinac, and Marcel Worring. 2019. HyperLearn: A Distributed Approach for Representation Learning in Datasets With Many Modalities. In *Proceedings of the 27th ACM International Conference on Multimedia*. 2245–2253.

[7] Tadas Baltrusaitis, Chaitanya Ahuja, and Louis-Philippe Morency. 2019. Multimodal Machine Learning: A Survey and Taxonomy. *IEEE Trans. Pattern Anal. Mach. Intell.* 41, 2 (Feb. 2019), 423–443. https://doi.org/10.1109/TPAMI.2018.2798607

[8] Ting Chen, Simon Kornblith, Mohammad Norouzi, and Geoffrey Hinton. 2020. A Simple Framework for Contrastive Learning of Visual Representations. arXiv:cs.LG/2002.05709

[9] Tat-Seng Chua, Jinhui Tang, Richang Hong, Haojie Li, Zhiping Luo, and Yantao Zheng. 2009. NUS-WIDE: A Real-World Web Image Database from National University of Singapore. In *Proceedings of the ACM International Conference on Image and Video Retrieval* (Santorini, Fira, Greece) *(CIVR '09)*. Association for Computing Machinery, New York, NY, USA, Article 48, 9 pages. https://doi.org/10.1145/1646396.1646452

[10] Keyan Ding, Ronggang Wang, and Shiqi Wang. 2019. Social Media Popularity Prediction: A Multiple Feature Fusion Approach with Deep Neural Networks. In *Proceedings of the 27th ACM International Conference on Multimedia* (Nice, France) *(MM '19)*. Association for Computing Machinery, New York, NY, USA, 2682–2686. https://doi.org/10.1145/3343031.3356062

[11] Fangxiang Feng, Xiaojie Wang, and Ruifan Li. 2014. Cross-Modal Retrieval with Correspondence Autoencoder. In *Proceedings of the 22nd ACM International Conference on Multimedia* (Orlando, Florida, USA) *(MM '14)*. Association for Computing Machinery, New York, NY, USA, 7–16. https://doi.org/10.1145/2647868.2654902

[12] Akash Gautam, Puneet Mathur, Rakesh Gosangi, Debanjan Mahata, Ramit Sawhney, and Rajiv Ratn Shah. 2019. #MeTooMA: Multi-Aspect Annotations of Tweets Related to the MeToo Movement. arXiv:cs.CL/1912.06927

[13] A. K. Gautam, L. Misra, A. Kumar, K. Misra, S. Aggarwal, and R. R. Shah. 2019. Multimodal Analysis of Disaster Tweets. In *2019 IEEE Fifth International Conference on Multimedia Big Data (BigMM)*. 94–103.

[14] Yunchao Gong, Qifa Ke, Michael Isard, and Svetlana Lazebnik. 2012. A Multi-View Embedding Space for Modeling Internet Images, Tags, and their Semantics. *CoRR* abs/1212.4522 (2012). arXiv:1212.4522 http://arxiv.org/abs/1212.4522

[15] Michael Gutmann and Aapo Hyvärinen. 2010. Noise-contrastive estimation: A new estimation principle for unnormalized statistical models. In *Proceedings of the Thirteenth International Conference on Artificial Intelligence and Statistics*. 297–304.

[16] Xin Hong, Pengfei Xiong, Renhe Ji, and Haoqiang Fan. 2019. Deep Fusion Network for Image Completion. In *Proceedings of the 27th ACM International Conference on Multimedia* (Nice, France) *(MM '19)*. Association for Computing Machinery, New York, NY, USA, 2033–2042. https://doi.org/10.1145/3343031.3351002

[17] Harold Hotelling. 1936. Relations Between Two Sets of Variates. *Biometrika* 28, 3/4 (1936), 321–377. http://www.jstor.org/stable/2333955

[18] Peng Hu, Liangli Zhen, Dezhong Peng, and Pei Liu. 2019. Scalable Deep Multimodal Learning for Cross-Modal Retrieval. In *Proceedings of the 42nd International ACM SIGIR Conference on Research and Development in Information Retrieval* (Paris, France) *(SIGIR'19)*. Association for Computing Machinery, New York, NY, USA, 635–644. https://doi.org/10.1145/3331184.3331213

[19] Olivier J. Hénaff, Aravind Srinivas, Jeffrey De Fauw, Ali Razavi, Carl Doersch, S. M. Ali Eslami, and Aaron van den Oord. 2019. Data-Efficient Image Recognition with Contrastive Predictive Coding. arXiv:cs.CV/1905.09272

[20] Armand Joulin, Edouard Grave, Piotr Bojanowski, and Tomas Mikolov. 2016. Bag of Tricks for Efficient Text Classification. *arXiv preprint arXiv:1607.01759* (2016).

[21] Onno Kampman, Elham J. Barezi, Dario Bertero, and Pascale Fung. 2018. Investigating Audio, Video, and Text Fusion Methods for End-to-End Automatic Personality Prediction. In *Proceedings of the 56th Annual Meeting of the Association for Computational Linguistics (Volume 2: Short Papers)*. Association for Computational Linguistics, Melbourne, Australia, 606–611. https://doi.org/10.18653/v1/P18-2096

[22] Meina Kan, Shiguang Shan, and Xilin Chen. 2016. Multi-view Deep Network for Cross-View Classification. *2016 IEEE Conference on Computer Vision and Pattern Recognition (CVPR)* (2016), 4847–4855.

[23] Dhruv Khattar, Jaipal Singh Goud, Manish Gupta, and Vasudeva Varma. 2019. MVAE: Multimodal Variational Autoencoder for Fake News Detection. In *The World Wide Web Conference* (San Francisco, CA, USA) *(WWW '19)*. Association for Computing Machinery, New York, NY, USA, 2915–2921. https://doi.org/10.1145/3308558.3313552

[24] Prannay Khosla, Piotr Teterwak, Chen Wang, Aaron Sarna, Yonglong Tian, Phillip Isola, Aaron Maschinot, Ce Liu, and Dilip Krishnan. 2020. Supervised Contrastive Learning. *arXiv preprint arXiv:2004.11362* (2020).

[25] Quoc V. Le and Tomas Mikolov. 2014. Distributed Representations of Sentences and Documents. *CoRR* abs/1405.4053 (2014). arXiv:1405.4053 http://arxiv.org/abs/1405.4053

[26] Tsung-Yi Lin, Michael Maire, Serge Belongie, James Hays, Pietro Perona, Deva Ramanan, Piotr Dollár, and C. Lawrence Zitnick. 2014. Microsoft COCO: Common Objects in Context. In *Computer Vision – ECCV 2014*, David Fleet, Tomas Pajdla, Bernt Schiele, and Tinne Tuytelaars (Eds.). Springer International Publishing, Cham, 740–755.

[27] Fei Liu, Jing Liu, Richang Hong, and Hanqing Lu. 2019. Erasing-based Attention Learning for Visual Question Answering. In *Proceedings of the 27th ACM International Conference on Multimedia*. 1175–1183.

[28] Laurens van der Maaten and Geoffrey Hinton. 2008. Visualizing data using t-SNE. *Journal of machine learning research* (2008).

[29] Sijie Mai, Haifeng Hu, and Songlong Xing. 2019. Modality to Modality Translation: An Adversarial Representation Learning and Graph Fusion Network for Multimodal Fusion. arXiv:cs.CV/1911.07848

[30] Niluthpol Chowdhury Mithun, Juncheng Li, Florian Metze, and Amit K. Roy-Chowdhury. 2018. Learning Joint Embedding with Multimodal Cues for Cross-Modal Video-Text Retrieval. In *Proceedings of the 2018 ACM on International Conference on Multimedia Retrieval* (Yokohama, Japan) *(ICMR '18)*. Association for Computing Machinery, New York, NY, USA, 19–27. https://doi.org/10.1145/3206025.3206064

[31] Niluthpol Chowdhury Mithun, Rameswar Panda, Evangelos E. Papalexakis, and Amit K. Roy-Chowdhury. 2018. Webly Supervised Joint Embedding for Cross-Modal Image-Text Retrieval. In *Proceedings of the 26th ACM International Conference on Multimedia* (Seoul, Republic of Korea) *(MM '18)*. Association for Computing Machinery, New York, NY, USA, 1856–1864. https://doi.org/10.1145/3240508.3240712

[32] Andriy Mnih and Koray Kavukcuoglu. 2013. Learning word embeddings efficiently with noise-contrastive estimation. In *Advances in neural information processing systems*. 2265–2273.

[33] Behnaz Nojavanasghari, Deepak Gopinath, Jayanth Koushik, Tadas Baltrušaitis, and Louis-Philippe Morency. 2016. Deep Multimodal Fusion for Persuasiveness Prediction. In *Proceedings of the 18th ACM International Conference on Multimodal Interaction* (Tokyo, Japan) *(ICMI '16)*. Association for Computing Machinery, New York, NY, USA, 284–288. https://doi.org/10.1145/2993148.2993176

[34] Ferda Ofli, Firoj Alam, and Muhammad Imran. 2020. Analysis of Social Media Data using Multimodal Deep Learning for Disaster Response. In *17th International Conference on Information Systems for Crisis Response and Management*. ISCRAM, ISCRAM.

[35] Liang Peng, Yang Yang, Zheng Wang, Xiao Wu, and Zi Huang. 2019. CRA-Net: Composed Relation Attention Network for Visual Question Answering. In *Proceedings of the 27th ACM International Conference on Multimedia*. 1202–1210.

[36] Yuxin Peng, Xin Huang, and Jinwei Qi. 2016. Cross-Media Shared Representation by Hierarchical Learning with Multiple Deep Networks. In *Proceedings of the Twenty-Fifth International Joint Conference on Artificial Intelligence* (New York, New York, USA) *(IJCAI'16)*. AAAI Press, 3846–3853.

[37] Yuxin Peng and Jinwei Qi. 2019. CM-GANs: Cross-Modal Generative Adversarial Networks for Common Representation Learning. *ACM Trans. Multimedia Comput. Commun. Appl.* 15, 1, Article 22 (Feb. 2019), 24 pages. https://doi.org/10.1145/3284750

[38] Yuxin Peng, Jinwei Qi, Xin Huang, and Yuxin Yuan. 2018. CCL: Cross-Modal Correlation Learning With Multigrained Fusion by Hierarchical Network. *Trans. Multi.* 20, 2 (Feb. 2018), 405–420. https://doi.org/10.1109/TMM.2017.2742704

[39] Y. Peng, X. Zhai, Y. Zhao, and X. Huang. 2016. Semi-Supervised Cross-Media Feature Learning With Unified Patch Graph Regularization. *IEEE Transactions on Circuits and Systems for Video Technology* 26, 3 (2016), 583–596.

[40] Hai Pham, Thomas Manzini, Paul Liang, and Barnabas Poczos. 2018. Seq2Seq2Sentiment: Multimodal Sequence to Sequence Models for Sentiment Analysis. 53–63. https://doi.org/10.18653/v1/W18-3308

[41] S. Poria, I. Chaturvedi, E. Cambria, and A. Hussain. 2016. Convolutional MKL Based Multimodal Emotion Recognition and Sentiment Analysis. In *2016 IEEE*

*16th International Conference on Data Mining (ICDM)*. 439–448.
[42] V. Ranjan, N. Rasiwasia, and C. V. Jawahar. 2015. Multi-label Cross-Modal Retrieval. In *2015 IEEE International Conference on Computer Vision (ICCV)*. 4094–4102.
[43] Jinfeng Rao, Hua He, and Jimmy Lin. 2016. Noise-contrastive estimation for answer selection with deep neural networks. In *Proceedings of the 25th ACM International on Conference on Information and Knowledge Management*. 1913–1916.
[44] Nikhil Rasiwasia, Jose Costa Pereira, Emanuele Coviello, Gabriel Doyle, Gert R.G. Lanckriet, Roger Levy, and Nuno Vasconcelos. 2010. A New Approach to Cross-Modal Multimedia Retrieval. In *Proceedings of the 18th ACM International Conference on Multimedia* (Firenze, Italy) *(MM '10)*. Association for Computing Machinery, New York, NY, USA, 251–260. https://doi.org/10.1145/1873951.1873987
[45] Nikhil Rasiwasia, Jose Costa Pereira, Emanuele Coviello, Gabriel Doyle, Gert R.G. Lanckriet, Roger Levy, and Nuno Vasconcelos. 2010. A New Approach to Cross-Modal Multimedia Retrieval. In *Proceedings of the 18th ACM International Conference on Multimedia* (Firenze, Italy) *(MM '10)*. Association for Computing Machinery, New York, NY, USA, 251–260. https://doi.org/10.1145/1873951.1873987
[46] Nikhil Rasiwasia, Jose Costa Pereira, Emanuele Coviello, Gabriel Doyle, Gert R.G. Lanckriet, Roger Levy, and Nuno Vasconcelos. 2010. A New Approach to Cross-Modal Multimedia Retrieval. In *Proceedings of the 18th ACM International Conference on Multimedia* (Firenze, Italy) *(MM '10)*. Association for Computing Machinery, New York, NY, USA, 251–260. https://doi.org/10.1145/1873951.1873987
[47] Jan Rupnik and John Shawe-Taylor. 2010. Multi-View Canonical Correlation Analysis. *SiKDD* (01 2010).
[48] Sebastian Schmiedeke, Pascal Kelm, and Thomas Sikora. 2012. Cross-Modal Categorisation of User-Generated Video Sequences. In *Proceedings of the 2nd ACM International Conference on Multimedia Retrieval* (Hong Kong, China) *(ICMR '12)*. Association for Computing Machinery, New York, NY, USA, Article 25, 8 pages. https://doi.org/10.1145/2324796.2324828
[49] Rajiv Shah and Roger Zimmermann. 2017. *Multimodal analysis of user-generated multimedia content*. Springer.
[50] Kai Shu. 2019. FakeNewsNet. https://doi.org/10.7910/DVN/UEMMHS
[51] Karen Simonyan and Andrew Zisserman. 2014. Very Deep Convolutional Networks for Large-Scale Image Recognition. arXiv:cs.CV/1409.1556
[52] Shivangi Singhal, Anubha Kabra, Mohit Sharma, Rajiv Ratn Shah, Tanmoy Chakraborty, and Ponnurangam Kumaraguru. 2020. SpotFake+: A Multimodal Framework for Fake News Detection via Transfer Learning (Student Abstract). (2020).
[53] S. Singhal, R. R. Shah, T. Chakraborty, P. Kumaraguru, and S. Satoh. 2019. SpotFake: A Multi-modal Framework for Fake News Detection. In *2019 IEEE Fifth International Conference on Multimedia Big Data (BigMM)*. 39–47.
[54] Kihyuk Sohn. 2016. Improved deep metric learning with multi-class n-pair loss objective. In *Advances in neural information processing systems*. 1857–1865.
[55] Christian Szegedy, Sergey Ioffe, Vincent Vanhoucke, and Alexander A. Alemi. 2017. Inception-v4, Inception-ResNet and the Impact of Residual Connections on Learning. In *Proceedings of the Thirty-First AAAI Conference on Artificial Intelligence* (San Francisco, California, USA) *(AAAI'17)*. AAAI Press, 4278–4284.
[56] Yonglong Tian, Dilip Krishnan, and Phillip Isola. 2019. Contrastive Multiview Coding. arXiv:cs.CV/1906.05849
[57] Yonglong Tian, Chen Sun, Ben Poole, Dilip Krishnan, Cordelia Schmid, and Phillip Isola. 2020. What makes for good views for contrastive learning. arXiv:cs.CV/2005.10243
[58] Thi Quynh Nhi Tran, Hervé Le Borgne, and Michel Crucianu. 2016. Cross-Modal Classification by Completing Unimodal Representations. In *Proceedings of the 2016 ACM Workshop on Vision and Language Integration Meets Multimedia Fusion* (Amsterdam, The Netherlands) *(iV&;L-MM '16)*. Association for Computing Machinery, New York, NY, USA, 17–25. https://doi.org/10.1145/2983563.2983570
[59] Aaron van den Oord, Yazhe Li, and Oriol Vinyals. 2018. Representation Learning with Contrastive Predictive Coding. arXiv:cs.LG/1807.03748
[60] Bokun Wang, Yang Yang, Xing Xu, Alan Hanjalic, and Heng Tao Shen. 2017. Adversarial Cross-Modal Retrieval. In *Proceedings of the 25th ACM International Conference on Multimedia* (Mountain View, California, USA) *(MM '17)*. Association for Computing Machinery, New York, NY, USA, 154–162. https://doi.org/10.1145/3123266.3123326
[61] Weiran Wang, Raman Arora, Karen Livescu, and Jeff Bilmes. 2015. On Deep Multi-View Representation Learning. In *Proceedings of the 32nd International Conference on Machine Learning - Volume 37* (Lille, France) *(ICML'15)*. JMLR.org, 1083–1092.
[62] Yaqing Wang, Fenglong Ma, Zhiwei Jin, Ye Yuan, Guangxu Xun, Kishlay Jha, Lu Su, and Jing Gao. 2018. EANN: Event Adversarial Neural Networks for Multi-Modal Fake News Detection. In *Proceedings of the 24th ACM SIGKDD International Conference on Knowledge Discovery & Data Mining* (London, United Kingdom) *(KDD '18)*. Association for Computing Machinery, New York, NY, USA, 849–857. https://doi.org/10.1145/3219819.3219903
[63] Zilong Wang, Zhaohong Wan, and Xiaojun Wan. 2020. TransModality: An End2End Fusion Method with Transformer for Multimodal Sentiment Analysis. In *Proceedings of The Web Conference 2020* (Taipei, Taiwan) *(WWW '20)*. Association for Computing Machinery, New York, NY, USA, 2514–2520. https://doi.org/10.1145/3366423.3380000
[64] Yiling Wu, Shuhui Wang, and Qingming Huang. 2018. Learning Semantic Structure-Preserved Embeddings for Cross-Modal Retrieval. In *Proceedings of the 26th ACM International Conference on Multimedia* (Seoul, Republic of Korea) *(MM '18)*. Association for Computing Machinery, New York, NY, USA, 825–833. https://doi.org/10.1145/3240508.3240521
[65] M. Wöllmer, F. Weninger, T. Knaup, B. Schuller, C. Sun, K. Sagae, and L. Morency. 2013. YouTube Movie Reviews: Sentiment Analysis in an Audio-Visual Context. *IEEE Intelligent Systems* 28, 3 (2013), 46–53.
[66] Xing Xu, Li He, Huimin Lu, Lianli Gao, and Yanli Ji. 2019. Deep Adversarial Metric Learning for Cross-Modal Retrieval. *World Wide Web* 22, 2 (March 2019), 657–672. https://doi.org/10.1007/s11280-018-0541-x
[67] Zhilin Yang, Zihang Dai, Yiming Yang, Jaime G. Carbonell, Ruslan Salakhutdinov, and Quoc V. Le. 2019. XLNet: Generalized Autoregressive Pretraining for Language Understanding. In *Advances in Neural Information Processing Systems 32: Annual Conference on Neural Information Processing Systems 2019, NeurIPS 2019, 8-14 December 2019, Vancouver, BC, Canada*, Hanna M. Wallach, Hugo Larochelle, Alina Beygelzimer, Florence d'Alché-Buc, Emily B. Fox, and Roman Garnett (Eds.). 5754–5764. http://papers.nips.cc/paper/8812-xlnet-generalized-autoregressive-pretraining-for-language-understanding
[68] Amir Zadeh, Minghai Chen, Soujanya Poria, Erik Cambria, and Louis-Philippe Morency. 2017. Tensor Fusion Network for Multimodal Sentiment Analysis. In *Proceedings of the 2017 Conference on Empirical Methods in Natural Language Processing*. Association for Computational Linguistics, Copenhagen, Denmark, 1103–1114. https://doi.org/10.18653/v1/D17-1115
[69] Xiaohua Zhai, Yuxin Peng, and Jianguo Xiao. 2014. Learning Cross-Media Joint Representation with Sparse and Semi-Supervised Regularization. *IEEE Transactions on Circuits and Systems for Video Technology* 24 (06 2014), 1–1. https://doi.org/10.1109/TCSVT.2013.2276704
[70] Xiaohua Zhai, Yuxin Peng, and Jianguo Xiao. 2014. Learning Cross-Media Joint Representation with Sparse and Semi-Supervised Regularization. *IEEE Transactions on Circuits and Systems for Video Technology* 24 (06 2014), 1–1. https://doi.org/10.1109/TCSVT.2013.2276704
[71] L. Zhang, B. Ma, G. Li, Q. Huang, and Q. Tian. 2018. Generalized Semi-supervised and Structured Subspace Learning for Cross-Modal Retrieval. *IEEE Transactions on Multimedia* 20, 1 (2018), 128–141.
[72] Liangli Zhen, Peng Hu, Xu Wang, and Dezhong Peng. 2019. Deep Supervised Cross-Modal Retrieval. In *The IEEE Conference on Computer Vision and Pattern Recognition (CVPR)*.

# Supplementary Material for COBRA: Contrastive Bi-Modal Representation Algorithm

## A  Overview

This document contains additional details for the paper **"COBRA: Contrastive Bi-Modal Representation Algorithm"**. Section B provides the description of the datasets used. It also mentions the number of training, validation and testing samples, and the features used. Section C describes the baselines used for comparison. Section D mentions the architectural details used for conducting the experiments. Section E contains the t-SNE plots of the joint embedding spaces learnt by our COBRA model for various datasets. It provides a visual confirmation of the existence of a highly discriminatory decision boundary.

## B  Dataset Description

In Table 8, we describe each dataset in detail, enumerating the number of classes, number of training, testing and validation samples, the features used for each modality and the task that the dataset was used for.

Table 8: Dataset Descriptions - */*/* in the samples column denotes the number of training/validation/test samples used. 'I', 'HC' and 'DS' for the CrisisMMD dataset refer to the 'Informativeness', 'Humanitarian Categories' and 'Disaster Severity' classication tasks respectively

| Dataset | Classes | Modality | Samples | Features | Task |
|---|---|---|---|---|---|
| Wikipedia | 10 | Image<br>Text | 2173/231/462<br>2173/231/462 | 4096-D VGG<br>300-D Doc2Vec | Cross-Modal Retrieval |
| PKU XMedia | 20 | Image<br>Text | 4000/500/500<br>4000/500/500 | 4096-D VGG<br>3000-D BoW | Cross-Modal Retrieval |
| NUS-Wide 10k | 10 | Image<br>Text | 8000/1000/1000<br>8000/1000/1000 | 4096-D VGG<br>1000-D BoW | Cross-Modal Retrieval |
| MS-COCO | 80 | Image<br>Text | 57455/14624/10000<br>57455/14624/10000 | 4096-D VGG<br>300-D Doc2Vec | Cross-Modal Retrieval |
| Politifact | 2 | Image<br>Text | 381/50/54<br>381/50/54 | 4096-D VGG<br>38400-D XLNet | Multi-Modal Fake News Detection |
| Gossipcop | 2 | Image<br>Text | 10010/1830/1000<br>10010/1830/1000 | 4096-D VGG<br>38400-D XLNet | Multi-Modal Fake News Detection |
| MeTooMA | 10 | Image<br>Text | 4500/1000/1000<br>4500/1000/1000 | 4096-D VGG<br>300-D Doc2Vec | Multi-Modal Sentiment Classification |
| CrisisMMD | I - 2<br>HC - 8<br>DS - 3 | Image<br>Text | 2000/793 (I&HC), 697 (DS)/500<br>2000/793 (I&HC), 697 (DS)/500 | 4096-D VGG<br>300-D Doc2Vec | Multi-Modal Disaster Classification |

## C  Baseline Models

To the best of our knowledge, we are the first to test a multi-modal classification model on the MeTooMA dataset. Hence, we evaluate our model against text-only uni-modal, image-only uni-modal, and early-fusion multi-modal baselines. For the baseline approaches, we use fully connected neural-networks for each of the 10 different classification tasks. Details about the baseline approaches are provided in Table 9.

Table 9: Baseline approaches for Multimodal Fine-grained Sentiment Classification

| Baseline Model | Description |
| --- | --- |
| Text-only baseline | Every text is vectorized using Doc2Vec embeddings and a fully connected network (with relu activations) is used with softmax outputs to perform the classification |
| Image-only baseline | Every image is vectorized using a pretrained VGG-19 network and a fully connected network (with relu activations) is used with softmax outputs to perform the classification |
| Early Fusion | Every image-text pair is vectorized (using the previously mentioned networks), concatenated directly, and a fully connected network (with relu activations) is used with softmax outputs to perform the classification |

## D  Architecture Details

For testing our COBRA model across different tasks, we ensure consistency for the encoder, decoder and classifier architectures. The architectural details are provided in Table 10.

Table 10: Architectures for encoder, decoder and classifer for all experiments. 'FC' denotes a fully connected layer, 'Concat' denotes feature-wise concatenation, 'ReLU' denotes the relu activation function and 'Dropout(p)' denotes a dropout layer with probability p.

| Encoder | Decoder | Classifier |
| --- | --- | --- |
| FC, $inp\_dims$, 1024, ReLU | FC, 512, 1024, ReLU | Concat($text\_feat$, $image\_feat$) |
| FC, 1024, 1024, ReLU | FC, 1024, 1024, ReLU | FC, $concat\_dims$, 512, ReLU, Dropout(0.5) |
| FC, 1024, 512, identity | FC, 1024, $inp\_dims$, identity | FC, 512, 128, ReLU, Dropout(0.5) |
| | | FC, 128, 64, ReLU, Dropout(0.2) |
| | | FC, 64, $num\_classes$ |

## E  t-SNE Plots

This section contains t-SNE plots of the joint embedding space representations obtained by COBRA on the different datasets. Figure 4 shows the 2-dimensional t-SNE plots for the XMedia and NUS-Wide 10k datasets. Figure 5 shows the 2-dimensional t-SNE plots for the different tasks of the CrisisMMD dataset. Figure 6 shows the 3-dimensional t-SNE plots for the Politifact, Gossipcop, Wikipedia, XMedia and NUS-Wide 10k datasets.

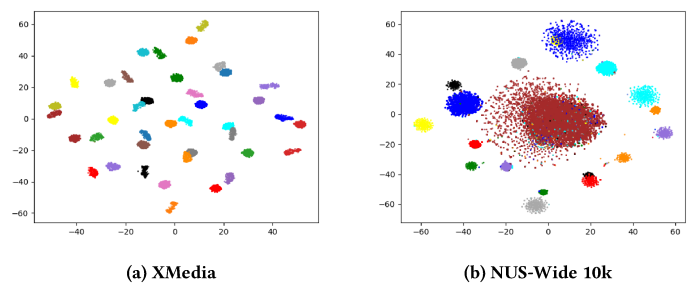

(a) XMedia    (b) NUS-Wide 10k

Figure 4: 2D t-SNE visualizations of the joint embedding spaces of the models trained on XMedia and NUS-Wide 10k datasets. The different colours correspond to the various class labels in the datasets. The circles represent text modality and the squares represent image modality.

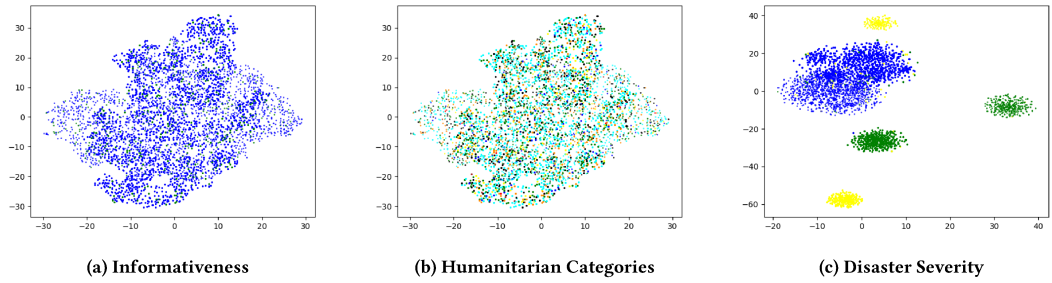

(a) Informativeness  (b) Humanitarian Categories  (c) Disaster Severity

Figure 5: 2D t-SNE visualizations of the joint embedding spaces of the models trained on different tasks of CrisisMMD dataset. The different colours correspond to the various class labels in the datasets. The circles represent text modality and the squares represent image modality.

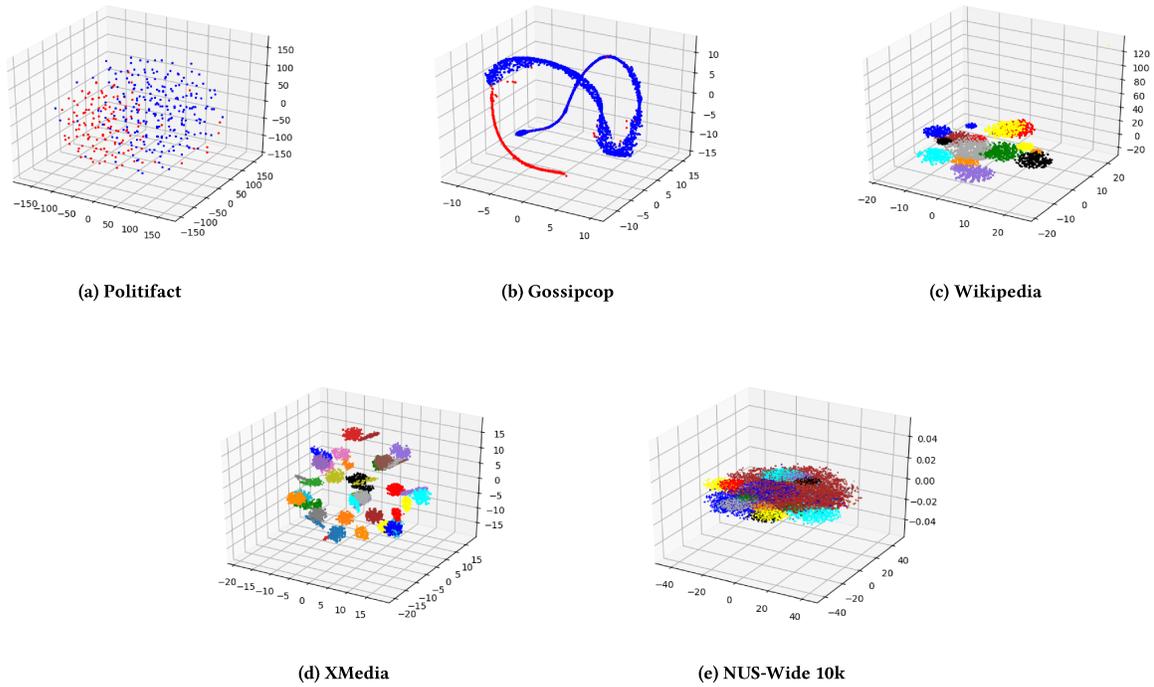

(a) Politifact  (b) Gossipcop  (c) Wikipedia

(d) XMedia  (e) NUS-Wide 10k

Figure 6: 3D t-SNE visualizations of the joint embedding spaces of the models trained on Politifact, Gosssipcop, Wikipedia, XMedia and NUS-Wide 10k datasets. The different colours correspond to the various class labels in the datasets. The circles represent text modality and the squares represent image modality.